\documentclass[letterpaper, 10 pt, conference]{ieeeconf}
\IEEEoverridecommandlockouts
\usepackage{graphics} 
\usepackage[line
snumbered,ruled]{algorithm2e}
\graphicspath{{img/}}
\usepackage{epsfig} 
\usepackage{subfigure}
\usepackage{amsmath} 
\usepackage{amssymb}  
\usepackage{amsthm}
\usepackage{bigints}
\usepackage{pifont}
\def\-{\raisebox{.75pt}{-}}
\usepackage{mathtools}
\usepackage{cite}
\usepackage[makeroom]{cancel}
\usepackage{xcolor} 

\usepackage{multirow}
\usepackage{hyperref}
\usepackage{booktabs}
\usepackage{epstopdf}
\usepackage{textcomp}
\usepackage{bm}

\usepackage{algpseudocode,algorithm} 
\usepackage{siunitx} 
\usepackage{color}
\usepackage{mathptmx}
\usepackage[11pt]{moresize}
\usepackage{hhline}
\usepackage{tikz}
\usepackage{textcomp}
\def\BibTeX{{\rm B\kern-.05em{\sc i\kern-.025em b}\kern-.08em
    T\kern-.1667em\lower.7ex\hbox{E}\kern-.125emX}}

\begin{document}
\title{\huge Inferring Spatial Uncertainty in Object Detection 
\thanks{$^{\ast}$ Zining Wang and Di Feng contributed equally to this work.}
\thanks{$^1$ Department of Mechanical Engineering, University of California, Berkeley, CA, 94720, USA.}
\thanks{$^2$ Robert Bosch GmbH, Corporate Research, Driver Assistance Systems and Automated Driving, 71272, Renningen, Germany.}
\thanks{$^3$ Institute of Measurement, Control and Microtechnology, Ulm University, 89081, Ulm, Germany.}
\thanks{Correspondence: \texttt{wangzining@berkeley.edu}. Video is available \url{https://youtu.be/mC7O3RZwhZc}}
}
\author{Zining Wang$^{\ast1}$, Di Feng$^{\ast2,3}$, Yiyang Zhou$^{1}$, Lars Rosenbaum$^{2}$, Fabian Timm$^{2}$, \\Klaus Dietmayer$^{3}$, Masayoshi Tomizuka$^{1}$, and Wei Zhan$^{1}$}

\maketitle

\begin{abstract}

The availability of real-world datasets is the prerequisite for developing object detection methods for autonomous driving. While ambiguity exists in object labels due to error-prone annotation process or sensor observation noises, current object detection datasets only provide deterministic annotations without considering their uncertainty. This precludes an in-depth evaluation among different object detection methods, especially for those that explicitly model predictive probability. In this work, we propose a generative model to estimate bounding box label uncertainties from LiDAR point clouds, and define a new representation of the probabilistic bounding box through spatial distribution. Comprehensive experiments show that the proposed model represents uncertainties commonly seen in driving scenarios. Based on the spatial distribution, we further propose an extension of IoU, called the Jaccard IoU (JIoU), as a new evaluation metric that incorporates label uncertainty. Experiments on the KITTI and the Waymo Open Datasets show that JIoU is superior to IoU when evaluating probabilistic object detectors.
\end{abstract}



\section{Introduction}\label{sec:introduction}
\begin{figure}[htpb]
	\centering
	\includegraphics[width=8cm]{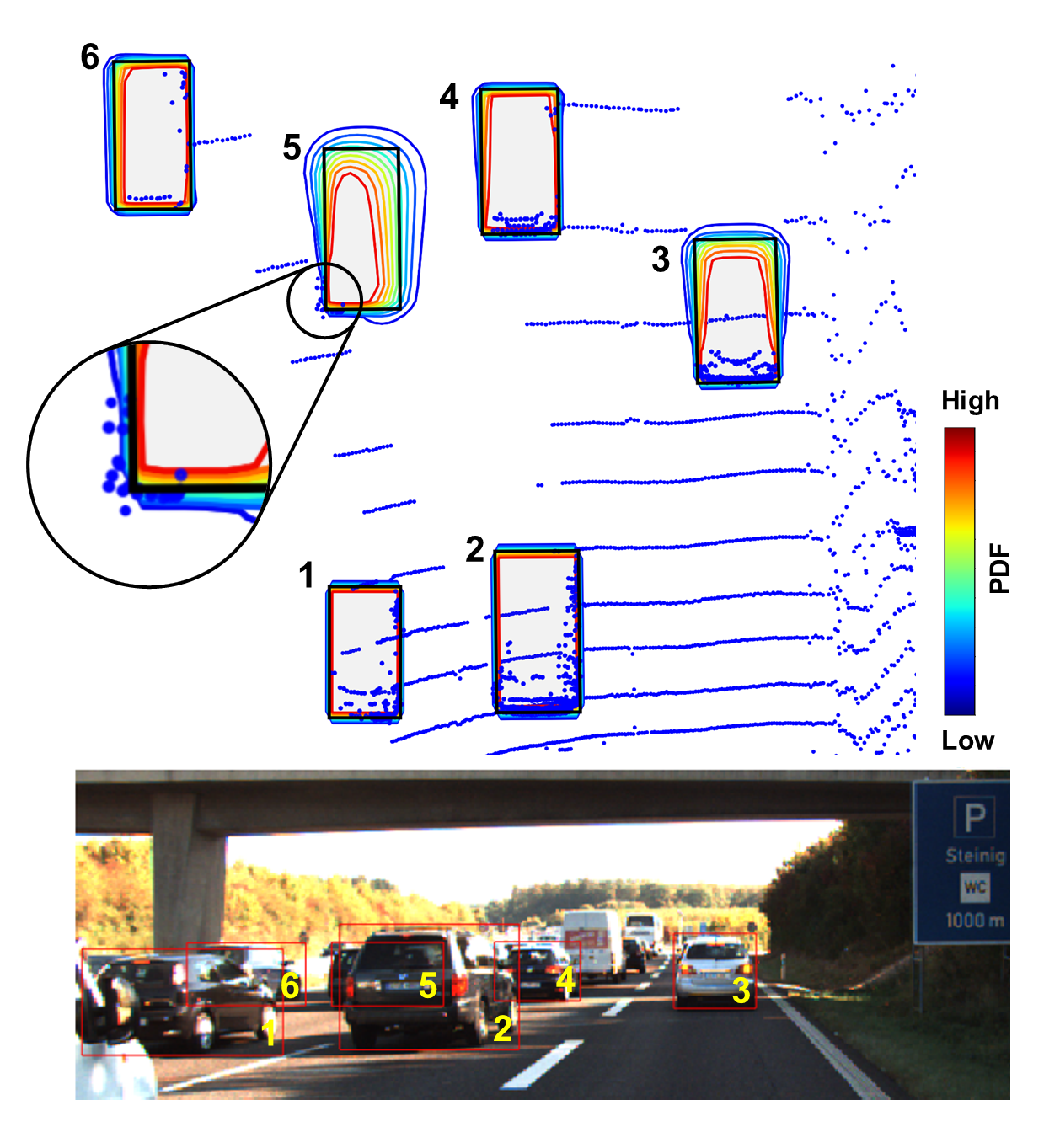}
	\caption{A demonstration of our proposed spatial uncertainty for bounding box labels in the KITTI dataset~\cite{geiger2012cvpr}. Objects are shown in the LiDAR Bird's Eye View (BEV). There exist errors (or uncertainty) inherent in labels. For object $3$, estimating its length is difficult because the surface information is only available on the side facing towards the ego-vehicle. For object $5$, the bounding box label does not even fully cover the LiDAR reflections near the bottom-left corer. Original data labels are deterministic, and they do not provide information on label wellness. In this work, we infer label uncertainty via a generative model of LiDAR points.} \label{fig:introduction}
	\vspace{-5mm}
\end{figure}
The availability of real-world driving datasets such as KITTI~\cite{geiger2012cvpr} and Waymo~\cite{sun2019scalability} is one of the key reasons behind the advancement of object detection algorithms in autonomous driving. However, the data labelling process can be error-prone due to human subjectivity and resource constraints. Ambiguity or uncertainty may also inherently exist in an object label. Think about the 3D object detection task, where annotators need to estimate the object positions only based on the surface information from cameras or LIDARs. Fig~\ref{fig:introduction} illustrates several bounding box labels of ``Car'' objects and their associated LiDAR point clouds in Bird's Eye View (BEV). The areas with dense LiDAR points (typically ``L-shape" areas) are easier to be labeled (e.g. object 2), whereas the back side of the object has higher labeling uncertainty due to insufficient observations (e.g. object 3). Ignoring such label uncertainty during training may degrade the generalization capability of an object detector since the model is forced to fit each training data sample equally, even the ones with remarkable noises. Significantly polluted data with noises can also deteriorate the detection performance~\cite{haase2019estimate}. Therefore, modelling label uncertainty in a dataset is indispensable for building robust, accurate object detectors for autonomous driving. Previous works have been focused on modelling class label noises in image classification problem~\cite{sukhbaatar2014training,lawrence2001estimating,xiao2015learning,vahdat2017toward}. To the best of our knowledge, modelling label uncertainties in object detection problem has not been widely studied, especially for bounding box labels. 

Uncertainties should be comprehensively considered, not only for the labels, but also for the evaluation metrics. Intersection over Union (IoU), defined as the geometric overlap ratio between two bounding boxes, is the most common metric to measure localization accuracy in object detection. Based on IoU, several metrics for detection accuracy are proposed, such as Average Precision (AP)~\cite{everingham2010pascal} and Localization Recall Precision (LRP)~\cite{oksuz2018localization}. However, those metrics are designed only for deterministic object detection: they can not be used to evaluate probabilistic object detectors~\cite{feng2018towards} which provide additional uncertainty estimation. The Probability-based Detection Quality (PDQ) metric~\cite{hall2018probabilistic} is designed specifically for probabilistic object detection. However, PDQ and the other aforementioned metrics only compare predictions with bounding box labels without considering the uncertainty (or ambiguity) in the labeling process. As a result, existing evaluation metrics may not fully reflect the performance of an object detector.

In this work, we explicitly model the uncertainty of bounding box parameters, which is inherent in labels (``label uncertainty'') for object detection datasets with LiDAR point clouds. The label uncertainty is inferred through a generative model of LiDAR points. In this way, we can easily incorporate prior knowledge of sensor observation noises and annotation ambiguity into our model. Then we propose the ``spatial distribution'' to visualize and represent the label uncertainty in 3D space or the LiDAR Bird's Eye View (BEV). We show that it reflects not only the typical L-shape observations in LiDAR point clouds, but also the quality of bounding box labels in a dataset. Based on the spatial distributions of bounding boxes, we further propose a probabilistic IoU, namely, Jaccard IoU (JIoU), as a new evaluation metric for object detection. The metric treats each bounding box label differently according to its uncertainty, and provides richer localization information than IoU. Using our proposed metric, we study the quality of uncertainty estimation from a state-of-the-art probabilistic object detector in KITTI~\cite{geiger2012cvpr} and Waymo~\cite{sun2019scalability} datasets.

In summary, our \textbf{contributions} are three-fold:
\begin{itemize}
    \item We infer the inherent uncertainty using a generative model in bounding box labels for object detection, and systematically analyze its parameters.
    \item We propose a new evaluation metric called JIoU for the object localization task, which considers label uncertainty and provides richer information than IoU when analyzing probabilistic object detectors. 
    \item We conduct comprehensive experiments with real-world datasets to justify the proposed method.
\end{itemize}
\section{Related Works}
\label{sec:related_works}
 
\subsection{Modelling Label Noises} \label{subsec:related_works:label_noises}
Explicitly modelling label noises (or uncertainty) has been an active research field~\cite{frenay2013classification}.
Whereas it is common to assume independent Gaussian noises for target regression, class label noises are much more complex to model, and incorrect assumptions may deteriorate the model performance (e.g. flipping class labels reverses the prediction results). Therefore, almost all proposed methods in the literature focus on modelling class label noises, especially in the image classification task~\cite{sukhbaatar2014training,lawrence2001estimating,xiao2015learning,vahdat2017toward}.
Only~\cite{meyer2019learning} and~\cite{meyer2019huber} are closely related to our work, which use simple heuristics to model uncertainties in bounding box labels for object detection. \cite{meyer2019learning} approximates the uncertainty with the IoU value between the label and the convex hull of aggregated LiDAR points. \cite{meyer2019huber} assumes Laplacian noises within bounding box labels. They interpret the Huber loss as the KL divergence between label uncertainty and predictive uncertainty and select hyper-parameters based on intuitive understanding of label uncertainties, in order to train object detection networks.  

Our methodology can be linked to the measurement models used in the LiDAR-based 3D single target object tracking~\cite{granstrom2011tracking,scheel2018tracking,hirscher2016tracking}. We assume a (generative) measurement model which generates noisy LiDAR measurements given a latent object state. However, instead of inferring a probability distribution over the latent object state from measurements over time and a prediction prior of a previous time-step, we infer a distribution over the latent object label from human annotators, given measurements of a single time-step and the assumption of known mean.

\subsection{Evaluation Metrics for Object Detection} \label{subsec:related_works:evaluation_metrics}
Intersection over Union (IoU) has been de fecto the standard metric to measure localization accuracy. It is often used to determine true positives and false positives among predictions, given a certain IoU threshold (e.g. in KITTI car detection benchmark, IoU is set to be $0.7$~\cite{geiger2012cvpr}). Furthermore, it has been extended as auxiliary losses during training to improve the detection performance~\cite{rezatofighi2019generalized,zhou2019iou,zheng2020distance,jiang2018acquisition}.
Based on the fixed IoU threshold, Average Precision (AP) is derived as the standard metric to measure detection accuracy~\cite{everingham2010pascal}. However, AP does not fully reflect the localization performance of a detection algorithm, since all predictions higher than the IoU threshold is treated equally. Observing this fact, the MS COCO benchmark~\cite{lin2014microsoft} calculates AP averaged over several IoU thresholds to show the difference. Oksuz \textit{et al.}~\cite{oksuz2018localization} propose a new evaluation metric called Localization Recall Precision (LRP) to incorporate the IoU score for each detection. The NuScenes object detection benchmark~\cite{caesar2019nuscenes} defines several new geometric metrics such as Average Translation Error (ATE), Average Scale Error (ASE) and Average Orientation Error (AOE) to specifically measure the bounding box localization performance. While those metrics only show the performance of deterministic object detection, Hall \textit{et al.}~\cite{hall2018probabilistic} focus on probabilistic object detection and propose the Probability-based Detection Quality (PDQ) metric, which jointly evaluates semantic probability (by the ``Label Quality" term) and spatial probability (by the ``Spatial Quality" term). The optimal PDQ is obtained when the predicted probability matches the absolute prediction error (e.g. a smaller IoU between a predicted bounding box and the ground truth indicates higher spatial uncertainty). 

\section{Methodology} \label{sec:methodology}
\subsection{Problem Formulation}
\label{sec:methodology:problem_formulation}
Labels in standard object detection datasets for autonomous driving (such as KITTI) usually include object classes $cls$, and deterministic object 3D poses and sizes $y$ in the form of bounding boxes. Denote $x_{all}$ as the set of all LiDAR points in a scan. Our target is to estimate the posterior distribution of bounding box labels, and extend IoU to a new metric called Jaccard IoU (JIoU) that incorporates spatial uncertainty. In this work, We demonstrate our method in vehicle objects.

We usually have the prior knowledge of object shape given its class $cls$. Therefore, it is possible to infer the posterior distribution of $y$ using the prior knowledge and observation $x_{all}$, i.e. $p(y|x_{all}, cls)$. For this purpose, we assume:
\begin{itemize}
    \item Labeling of class $cls$ is accurate.
    \item Segmentation is accurate. The set of points $x$ belonging to the object segmented out from $x_{all}$ by human annotation has few outliers.
    \item Human-labeled bounding box parameters $\overline{y}$ is the accurate mean of $y$. We only care about the spread of $y$.
\end{itemize}

Under these assumptions, the position, rotation and size of an object $y$ is only conditioned on the observation of points belonging to this object, denoted as $x=\{x_1,\cdots,x_K\}$. The posterior distribution $p(y|x_{all}, cls)$ is denoted as $p(y|x)$ for notation simplicity. We are especially interested in $p(y|x)$ because the LiDAR observations provided to the human annotators are usually not enough for determining the full size and center of the object, due to occlusion and sensor noises.

\begin{figure}[h]
	\centering
	\vspace{-3mm}
	\begin{minipage}{0.45\linewidth}
		\centering
		\includegraphics[width=0.85\linewidth]{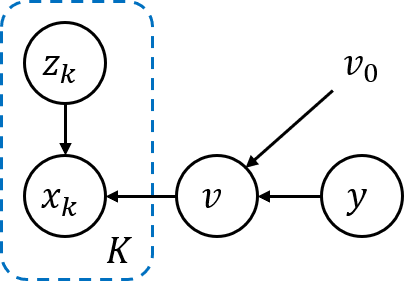}
	\end{minipage}
	\hskip 2pt
	\begin{minipage}{0.4\linewidth}
		\centering
		\caption{Probabilistic graphical model for the inference of $y$. $z_k$ is the latent variable associated with each observation, which contains the semantic meaning such as the part of the object that the point $x_k$ belongs to.} \label{fig:PGM}
	\end{minipage}
	\vspace{-3mm}
\end{figure}

An object in detection task is parameterized by its center location $c_1,c_2,c_3$, 3D extents $l,w,h$ and orientation $r_y$, i.e. $y=[c_1,c_2,c_3,l,w,h,r_y]$. This provides a universal affine transformation from the point $v_0\in[-0.5,0.5]^3$ of the unit bounding box $B(y^*)$ to the actual point $v(y)$ on the object surface in the 3D space:
\begin{equation}
    \label{eq:zy_render}
    v(y) = R_y
\left[\begin{smallmatrix}
l& 0& 0\\
0& w& 0\\
0& 0& h
\end{smallmatrix}\right]
v_0 + \left[\begin{smallmatrix}
c_1\\
c_2\\
c_3
\end{smallmatrix}\right],
\end{equation}
with $R_y$ being the rotation matrix from $r_y$, and $v_0$ depending on the prior knowledge of the shape, e.g. from the CAD model, bounding box or point cloud rendering method~\cite{fang2018simulating}. The graphical model is then illustrated in Fig.~\ref{fig:PGM}. In this paper, we uniformly sample $v_0$'s from the bounding box boundary to generate the object surface for simplicity.

In the sequal, Section~\ref{sec:methodology:label_uncertainty} introduces how we derive the posterior distribution $p(y|x)$ via a variational Bayes method. The derived distribution $p(y|x)$ is in the parameter space of $y\in \mathbb{R}^7$, which is difficult to visualize and represent the uncertainty of a bounding box. Therefore, Section~\ref{sec:uncertainty} proposes to transform the distribution $p(y|x)$ into a distribution $p(u)$ in the 3D space $u \in \mathbb{R}^3$ or BEV space $u\in \mathbb{R}^2$. Such label spatial distribution allows us to extend IoU to a metric (JIoU) that measures the probabilistic object localization performance, described in Section~\ref{sec:methodology:JIoU}.

\subsection{Label Uncertainty from Generative Model of Point Cloud}
\label{sec:methodology:label_uncertainty}

\begin{figure}[h!]
	\centering
	\vspace{-3mm}
	\includegraphics[width=4.5cm]{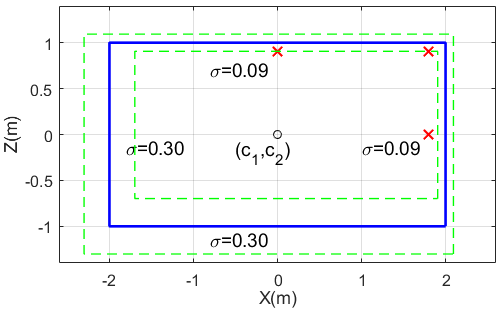}
	\vspace{-3mm}
	\caption{A simple demonstration of calculating the posterior with deterministic latent variable. $x_1{=}(1.8,0),\ x_2{=}(1.8,0.9),\ x_3{=}(0,0.9)$ and corresponding $v$'s are $v_1{=}(c_1{+}0.5l,0),\ v_2{=}(c_1{+}0.5l,c_2{+}0.5w),\ v_3{=}(0,c_2{+}0.5w)$} \label{fig:demo1}
	\vspace{-3mm}
\end{figure}

We start to elaborate our method with a simple example given in Fig.~\ref{fig:demo1} in the bird's eye view (BEV) with point clouds as observation $x = \{x_1,x_2,x_3\}$. Three points are segmented out inside the bounding box as red cross markers. The posterior is solved by Bayes rule:
\begin{equation} \label{eq:bayes1}
    p(y|x) = \frac{p(x|y)p(y)}{p(x)},
\end{equation}
assuming that $p(x|y)=\prod_{k=1}^{K}p(x_k|y;v_k)$, $K{=}3$, where each point $x_k\in\mathbb{R}^2$ is independently generated by the nearest point $v_k(y)\in\mathbb{R}^2$ on the boundary of the ground truth bounding box with Gaussian noise:
\[p(x_k|y;v_k)\sim \mathcal{N}\left(v_k(y),\sigma^2\right),\quad \sigma{=}0.2m\]
\par The label parameters are center and size $y{=}[c_1,c_2,l,w]$. The posterior $p(y|x)$ is then Gaussian given Gaussian prior $p(y)\sim\mathcal{N}([0,0,4,2],100^2)$:
\[p(y|x)=\prod_{k=1}^{3}{p(x_k|y;v_k)}p(y)\sim \mathcal{N}(\cdot,0.01\times
\left[\begin{smallmatrix}
4& 0& {-}4& 0\\
0& 4& 0& {-}4\\
-4& 0& 6& 0\\
0& -4& 0& 6
\end{smallmatrix}\right]).\]

Different from others~\cite{meyer2019learning,feng2018leveraging} who calculate the uncertainty of predictions and labels as independent variables, the uncertainty derived here is a joint distribution with correlation. A singular value decomposition (SVD) of the covariance matrix shows that two edges of the bounding box which have point cloud observations, namely $X=c_1+l/2$ and $Z=c_2+w/2$, have the smallest standard deviations $0.09m$ and $0.09m$. The other two edges without observation have the largest standard deviations $0.43m$ and $0.43m$. Fig.~\ref{fig:demo1} illustrates the confidence interval within one standard deviation by green dashed bounding boxes. The advantage of producing a joint distribution is significant. For those who produce disjoint distributions on size and pose, the confidence intervals or variances are the same between face and back, and between left and right side of the car. Meanwhile, it is the common sense that surfaces with more observations should have less variance, as shown by Fig.~\ref{fig:demo1}.\par

A more general model is by assuming that $p(x|y)$ is some mixture model with categorical latent variable $z$, e.g. a Gaussian Mixture Model (GMM) of center points $v_j(y), j{=}1,2,\cdots,M$: 
\begin{equation} \label{eq:GMM_generative_model}
    p(x_k|y) = \sum_{j=1}^{M}p\left(z_k{=}j\right)\mathcal{N}\left(v_j(y),\sigma_j^2I\right),
\end{equation}
Each $v_j(y)$ is created from a unique point $v_{j0}$ inside the unit bounding box $B(y^*):{=} [-0.5,0.5]^3$, by Eq.~\ref{eq:zy_render}. $\sigma_j^2$ is an empirical covariance related to the sensor noise and the confidence of the rendering method. We use the surface of boundary of the box to create $v_0$ and $\sigma_j{=}\sigma$ for all $j$, but it does not exclude the potential of more complex rendering methods. 

An approach to solve the posterior of GMM is by variational Bayes (VB), assuming $y,z$ are independent. A good practice of VB is the point registration method. The problem then becomes solving $q(y)$, $q(z)$ that minimize the KL-divergence between the assumed class of distribution and the actual posterior:
\begin{equation} \label{eq:VB_problem}
    D_{KL}\left(q||p\right) = \int_{z} q(y,z)\log{\frac{q(y,z)}{p(y,z|x)}},
\end{equation}
and the solution is given below by mean field method~\cite{blei2017variational}:
\begin{equation} \label{eq:VB_solution}
\begin{aligned}
    q(y)&\propto exp{\Big\{\log{p(y)}+\mathbb{E}_{z}\left[\log{p(x|z,y)}\right]\Big\}}\\
        &\propto exp\bigg\{-\sum_{j=1}^{M}\frac{1}{2\sigma_j^2}{\sum_{k=1}^{K}{\varphi_{jk}||x_k-v_j(y)||^2}}\bigg\},
\end{aligned}
\end{equation}
with $\varphi_{jk}:=p(z_k=j|x_k)$ being the probability of registering $x_k$ to $v_j$ and it can be calculated using the nominal value $\overline{y}$ of the ground truth bounding box:
\begin{equation}
    \label{eq:registrator}
    \varphi_{jk} = \frac{\exp\left(-\frac{1}{2\sigma_j^2}||x_k-v_j(\overline{y})||^2\right)}{\sum_{j=1}^M{\exp\left(-\frac{1}{2\sigma_j^2}||x_k-v_j(\overline{y})||^2\right)}},
\end{equation}
where $v_j(y)$ is a linear function of $y$ in our example used for demonstration and the resulting posterior $p(y|x)$ is Gaussian 

The proposed variational Bayes method Eq.~\ref{eq:VB_solution} is used to calculate the label uncertainty of vehicles in terms of detection based on point cloud. The generative model is generated by the ground truth bounding box. For extensions to pedestrians and other objects, a more dedicated rendering model is desired and it is left for future development.

\subsection{Spatial Distribution for Probabilistic Bounding Box}
\label{sec:uncertainty}
Regardless the uncertainty of labels, a proper representation of the probabilistic bounding box is desired for evaluating the probabilistic detection ~\cite{feng2019can,hall2018probabilistic, pan2020towards}. We propose a generative model that generates a spatial distribution of the bounding box in 3D or BEV. It provides a visualization of uncertainty and is later shown that it supports the extension of the commonly used IoU.

The idea of probabilistic box representation is proposed in PDQ~\cite{hall2018probabilistic} for 2D axis-aligned bounding box of image. The resulting spatial distribution $P(u)\in \left[0,1\right]$ denotes the probability of pixel $u\in \mathbb{Z}^2$ belonging to the object. A natural generalization of $P(u)$ to $u\in \mathbb{R}^3$ in 3D space or $u\in\mathbb{R}^2$ in BEV for rotated bounding box $B(y)$ is: 
\begin{equation}
    \label{eq:Probabilistic_object}
    P_{PDQ}(u) := \int_{\{y|u\in B(y)\}}{p_{\widehat{Y}}(y|x)dy},
\end{equation}
where $P(u)$ is the probability that $u$ is a point of the object. The subscript $\widehat{Y}$ (or $Y$) is the random variable of the detection (label). Eq.~\ref{eq:Probabilistic_object} is easy to calculate for axis-aligned bounding boxes but hard for rotated boxes because it has to integrate over the space of $y$, which is 7 dimensional for 3D. A transformation in the integral gives another expression as a probabilistic density fumction (PDF): 
\begin{equation}
    \label{eq:Generative_object}
    \begin{aligned}
    p_G(u)&:=\int_{v_0\in B(y^*)}p_{V\left(v_0,\widehat{Y}\right)}\left(u\right)dv_0\\
    &=\int_{\{y|u\in B(y)\}}{\frac{1}{A(y)} p_{\widehat{Y}}(y|x)dy},
    \end{aligned}
\end{equation}
where $V(v_0,Y)$ is defined in Eq.~\ref{eq:zy_render}, and $V$ and $Y$ are random variables. $v_0$ is added as an argument because we need to integrate over $v_0$. Given the probabilistic density of $Y$, e.g. from Section~\ref{sec:methodology:label_uncertainty}, it is not difficult to get the density of $V(v_0,Y)$ as Eq.~\ref{eq:zy_render} is quite simple. 

The proposed definition of spatial distribution $p_G$ is slightly different from $P_{PDQ}$ by scaling the density with the size $A(y)$ of the bounding box $B(y)$. The scaling factor enables the transformation to a integral over distributions generated by points $v_0$ inside the unit box $B(y^*)$, i.e. $A\left(y^*\right){=}1$. Therefore, it has a significant advantage of reducing the dimension of the integral from 7 to 3 for 3D. Besides, $\int_u p_G(u){=}1$ if it is integrated on spatial points $u$, but $P_{PDQ}(u)$ is not normalized. The shapes of their distribution differ only when the object size is uncertain and the proposed $p_G(u)$ tend to be more concentrated.


\subsection{JIoU: A Generalization of IoU for Evaluation with Uncertainty}
\label{sec:methodology:JIoU}
IoU is one of the most commonly used metric for detection. It has an intuitive geometry definition measuring the overlap between the predicted and ground truth bounding boxes. Despite its popularity, IoU only applies to deterministic predictions and labels. In this section, we define a metric over $p(u)$ of probabilistic bounding box following the definition of the probabilistic Jaccard index~\cite{moulton2018maximally}:
\begin{equation}
    \label{eq:JIoU_def}
    JIoU := \int_{R_{1}\cap R_{2}} {\frac{du}{\int_{R_1 \cup R_{2}}{\max\left(\frac{p_1(v)}{p_1(u)},\frac{p_2(v)}{p_2(u)}\right)dv}}},
\end{equation}
$p_1,p_2$ are the spatial distributions of two boxes, as introduced in Eq.~\ref{eq:Generative_object}. $u, v$ are points in the 3D or BEV space. $R_1,R_2$ are the supports of $p_1,p_2$, respectively. Note that JIoU degenerates to IoU when two boxes are deterministic, i.e., $p(y|x)$ is a delta function, where $R_1, R_2$ become bounding boxes and $p_1, p_2$ become uniform inside their boxes. 
Some properties of JIoU can be concluded as:
\begin{itemize}
    \item $JIoU=IoU$ if two boxes are both deterministic 
    \item The computational complexity is $O(NlogN)$ using Eq. 3 of~\cite{moulton2018maximally} if $N$ points are sampled from $R_1\cup R_2$.
\end{itemize}

\begin{figure}[h!]
	\centering
	\subfigure[$\text{JIoU}{=}0.1$ using $P_{PDQ}$]{\includegraphics[width=3cm]{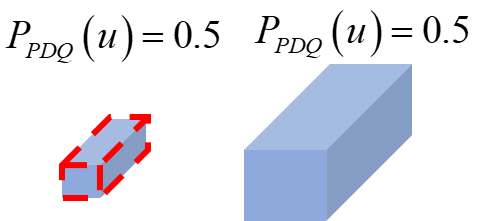}}
	\hspace{1mm}
	\subfigure[$\text{JIoU}{=}0.5$ using $p_{G}$]{\includegraphics[width=2.8cm]{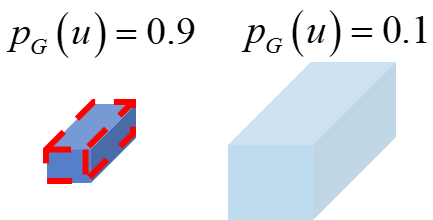}}
	\caption{Spatial distributions of discrete $Y$ with two possible values in blue and predicted box $\hat{Y}=\hat{y}$ in dashed red line.}
	\vspace{-4mm}
	\label{fig:JIoU_with_different_distribution}
\end{figure}

The proposed $p_G(u)$ is more reasonable than $P_{PDQ}(u)$ when JIoU is used as a metric. Consider when label $Y$ is a discrete random variable with two values of equal probability, i.e. two possible bounding boxes, and when the prediction $\hat{Y}$ only fits one of the box as shown in Fig.~\ref{fig:JIoU_with_different_distribution}. Then $P_{PDQ}(u)=0.5$ for the label and $JIoU(Y,\hat{Y}){\approx} 0$ because one box is much smaller. On the contrary, $JIoU(Y,\hat{Y}){=}0.5$ under $p_G$, no matter what the size difference is. Here it is desired that $JIoU=0.5$ because the prediction has matched one of the two possible ground truths. 
\section{Experimental Results} \label{sec:experimental_result}
In this work, we propose (1). a generative model to infer the uncertainty of bounding box labels for LiDAR point clouds, (2). a new spatial distribution to visualize the uncertainty, and (3). JIoU as an extension of IoU to evaluate probabilistic object detectors. In the following, we design three experiments to verify our proposed methods. First, we study how the model parameters affect the label uncertainty, including the LiDAR observation noises and the prior knowledge of human annotators (Sec.~\ref{sec:experimental_result:choice_of_parameters}). Second, we justify our methods and show that the spatial uncertainty reflects the typical ``L''-shape behaviours in LiDAR point clouds, which are also observed in state-of-the-art object detection networks (Sec.~\ref{sec:experimental_result:uncertainty_justification}). In addition, we show our methods reflect the quality of bounding box labels (Sec.~\ref{sec:experimental_result:label_wellness}). Third, we use JIoU as a new evaluation metric to explore predictions from probabilistic object detection networks (Sec.~\ref{sec:experimental_result:jiou}).

We evaluate the proposed method on the KITTI dataset~\cite{geiger2012cvpr} and the recently released Waymo open dataset~\cite{sun2019scalability}. Both provide LiDAR observations and 3D object labels from human annotators. We use the ``Car'' category on the KITTI training dataset, with $7481$ data frames and nearly $30$K objects. As for the Waymo dataset, we select the training data drives recorded in San Francisco, and down-sample the frames by a factor of $10$. The original ``Vehicle'' class in the Waymo dataset does not distinguish among objects such as motorcycles, cars or trucks, making it difficult to directly compare with the KITTI dataset. Therefore, we extract the ``Car'' objects from the ``Vehicle'' class by thresholding the vehicle length within $3$m$-6.5$m. We use such modified Waymo database with $7545$ frames and over $150$K objects. When evaluating our proposed label uncertainty with the help of object detection networks, We report the standard Average Precision (AP) from the BEV ($AP_{BEV}$), with IoU=0.7 threshold. Objects in the KITTI dataset are categorized into ``Easy'', ``Moderate'' and ``Hard'' settings~\cite{geiger2012cvpr}, while objects in the Waymo dataset are categorized by thresholding the LiDAR range up to 30m, 50m and 70m.

\subsection{Choice of Parameters for Label Uncertainty}\label{sec:experimental_result:choice_of_parameters}
\begin{figure}[h!]
	\centering
	\vspace{-5mm}
	\includegraphics[width=8.5cm]{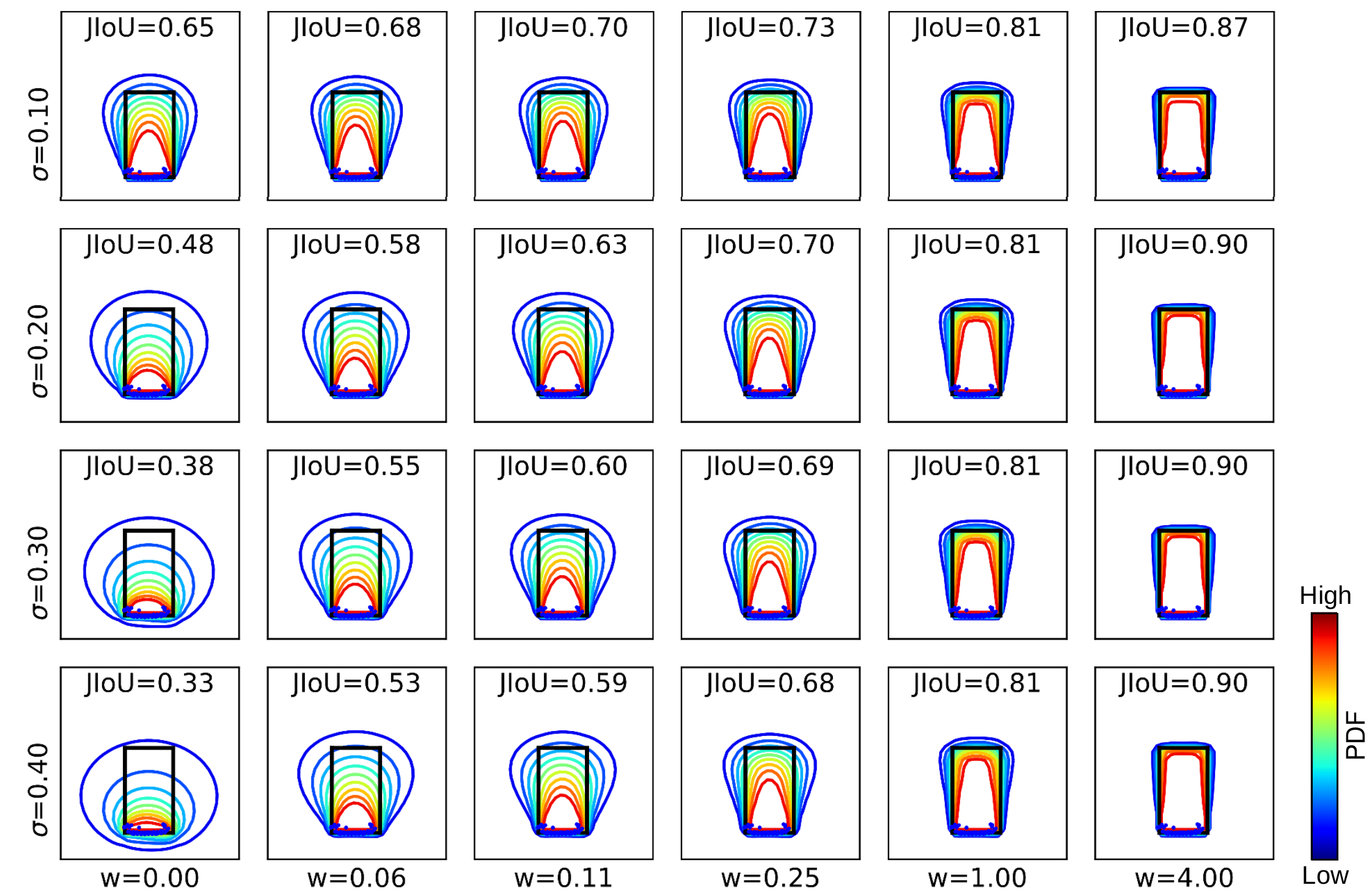}
	\caption{Influence of LiDAR measurement noise and prior distribution on spatial distribution. The sample is drawn from frame 1287, object 11 in KITTI.} \label{fig:exp_parameter_influence}
	\vspace{-3mm}
\end{figure}

The inference of label uncertainty $p(y|x)$ introduced in section~\ref{sec:methodology:label_uncertainty} allows us to incorporate prior knowledge of LiDAR observation noises $p(x|y,z)$. Furthermore, it can incorporate models for human annotation uncertainty in the priors $p(y)$. This section explores the impact of parameters on the label uncertainty. More specifically, we examine the standard deviation of LiDAR observation noise $\sigma$, and the covariance of $p(y)$. We empirically define the covariance matrix in BEV of $y=[c_1,c_2,l,w,r_y]^T$, referring to the variance of vehicle size for all objects in the KITTI dataset: $1/w \times \text{diag}([0.44^2, 0.11^2, 0.25^2, 0.25^2, 0.17^2])$,
where $w$ is the weight of the prior knowledge. Larger $w$ means smaller variance of prior distribution and more confidence in human annotators. Fig.~\ref{fig:exp_parameter_influence} shows the evolution of spatial distribution $p(u)$ and JIoU score for a bounding box label with increasing $\sigma$ and $w$. There are only a few LiDAR observations on the front surface of the vehicle. This results in high uncertainty, or low density, at the opposite side of the bounding box label, if no prior distribution is added. As more prior knowledge is incorporated (e.g.. increasing weights $w$ for human annotations, or decreasing observation noise $\sigma$), the posterior uncertainty decreases, resulting in higher JIoU score. Furthermore, we observe that labels with lots of observed LiDAR points are almost never affected by the choice of parameters. They have an uniform spatial distribution inside bounding boxes, even without prior distribution.

Instead of empirically choosing the value of $\sigma$, it can be estimated by the EM algorithm in~\cite{lawrence2001estimating} using
\begin{equation}
    \label{eq:parameter_EM}
    \sigma^2=\frac{1}{Md}\sum_{j=1}^{M}{\sum_{k=1}^{K}{\varphi_{jk}||x_k-z_j(\overline{y})||^2}},
\end{equation}
with iterative updates. The resulting $\sigma$ is $0.2$m for KITTI and $0.3$m for Waymo. Eq.~\ref{eq:parameter_EM} implies that these results are very close to the root mean square of distances between LiDAR points and bounding box labels. The chosen value of $\sigma^2$ includes both the measurement noise of the LiDAR sensor and the approximation error of a bounding box to the actual surface of a car.

\subsection{Justification of uncertainty model}\label{sec:experimental_result:uncertainty_justification}
\begin{figure}[tpb]
	\centering
	\begin{minipage}{0.55\linewidth}
		\centering
		\includegraphics[width=1\linewidth]{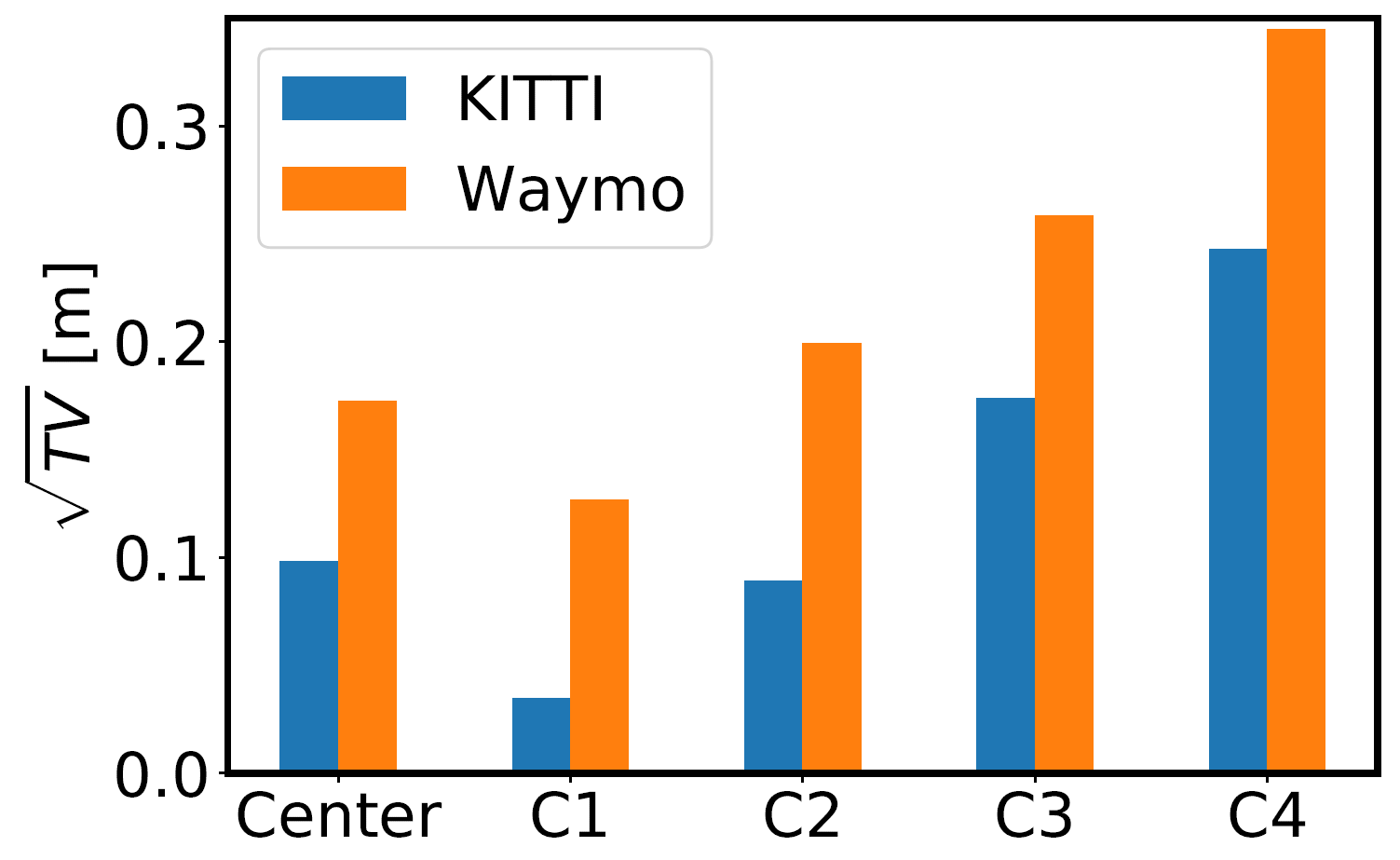}
	\end{minipage}
	\hskip 2pt
	\begin{minipage}{0.4\linewidth}
		\centering
		\caption{The average total variances~\cite{feng2018leveraging} of centers and corners of all ``Car'' objects in the KITTI dataset. Variances are calculated by the proposed uncertainty model. Corners (C1-C4) are sorted by their distances to the ego-vehilce.} \label{fig:corner_TVs}
	\end{minipage}
	\vspace{-3mm}
\end{figure}

The uncertainty model proposed in this paper suggests that different points of the bounding box have different variances: points close to the LIDAR are likely to have smaller variances. The importance of modeling the variance of different points is also revealed in a probabilistic detection paper ~\cite{pan2020towards}.  Fig.~\ref{fig:corner_TVs} measures the average total variances (TV) of each of the four corners in the BEV bounding boxes in the KITTI and Waymo datasets (C1-C4, sorted by their distance to the ego-vehicle in the ascending order). The TV scores are calculated by our proposed spatial distribution. We observe that the nearest corner C1, which usually has dense laser points are more reliable than the center of the box and the distant, occluded corders, by showing smaller TV scores. This observation corresponds to the intuition of L-shapes~\cite{zhang2017efficient} widely used for vehicle detection and tracking. The nearest corner of the L-shape is determined more confidently by our method while the size of the vehicle is more uncertain. 
Considering the mechanism of LiDAR perception, we expect that using object detectors to predict the nearest corner for location and orientation should be less prone to label noise than predicting the center. Table~\ref{table:L-Shape} verifies this result and shows how a small modification of ``corner'' and ``center'' directly benefits the AP of multiple networks.

In the table, We modify the outputs from several state-of-the-art LiDAR-based object detectors, and show how the location and orientation accuracy is governed by L-shape. We use STD~\cite{yang2019std}, SECOND~\cite{yan2018second}, AVOD~\cite{ku2017joint}, VoxelNet~\cite{zhou2017voxelnet} and PointRCNN~\cite{Shi_2019_CVPR} in the KITTI dataset, and PIXOR~\cite{yang2018pixor} in the Waymo dataset. Following the nuScenes~\cite{caesar2019nuscenes} evaluation criterion which uses partial ground truth information to adjust predictions in order to separate location and size accuracy, we evaluate networks by replacing size predictions with ground truth labels, while keeping the center predictions or the nearest corner predictions unchanged. 

\begin{figure}[tpb]
	\centering
	\begin{minipage}{0.55\linewidth}
		\centering
		\includegraphics[width=1\linewidth]{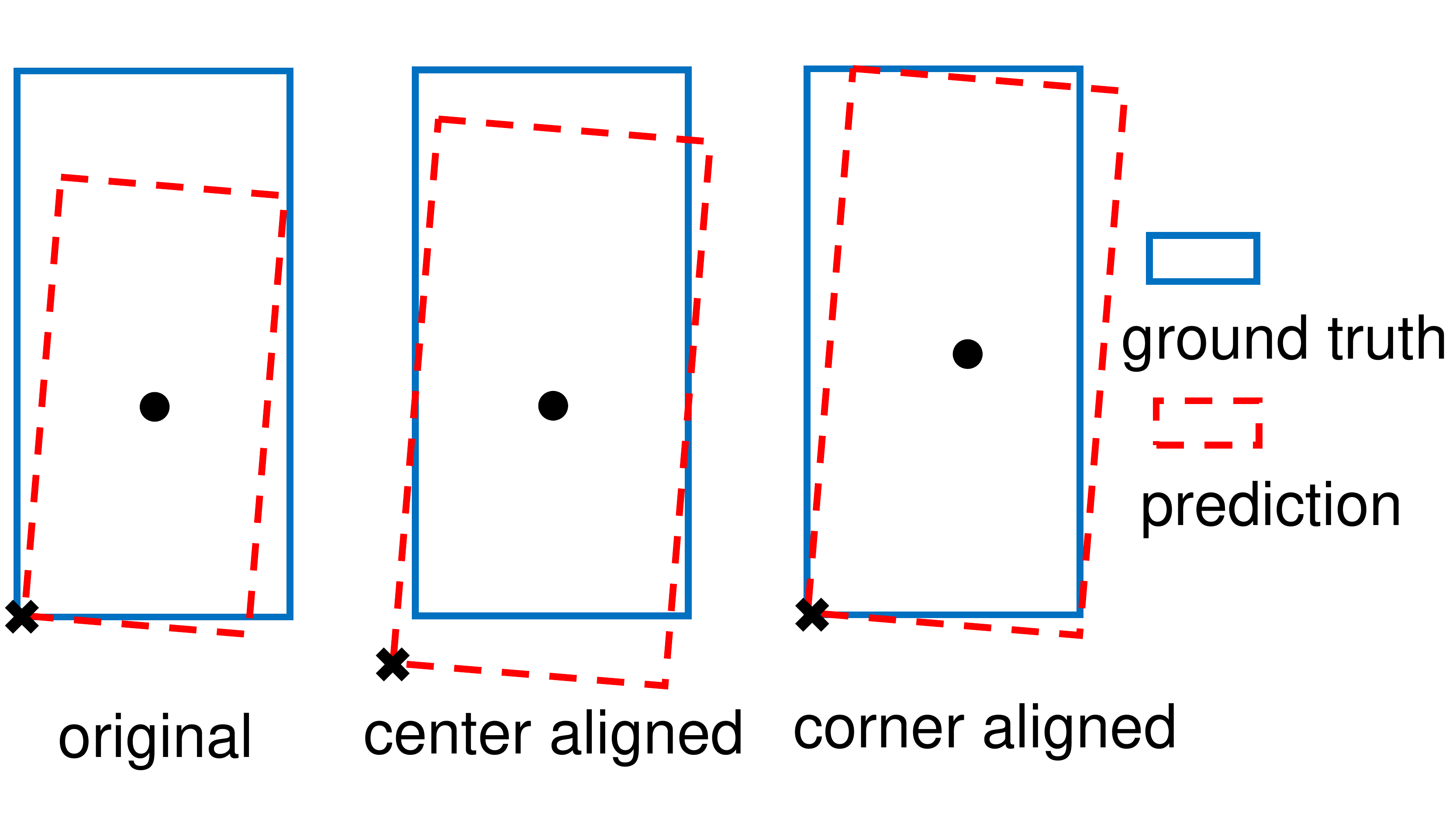}
	\end{minipage}
	\hskip 2pt
	\begin{minipage}{0.4\linewidth}
		\centering
		\caption{Column ``origin'' in Table~\ref{table:L-Shape} does no modification to the predicted bounding box. ``center aligned'' assembles the bounding boxes using center predictions and ground truth sizes. ``corner aligned'' using nearest corner predictions and ground truth sizes. } \label{fig:Lshapes}
	\end{minipage}
	\vspace{-3mm}
\end{figure}

The predicted bounding boxes are reassembled as shown in Fig~\ref{fig:Lshapes}. ``corner aligned'' consistently demonstrates much better $AP_{BEV}$ improvement than ``center aligned'' among all networks and datasets. The result justifies that the networks have better predictions on corners than centers. Our proposed label uncertainty naturally captures this behaviour, and has the potential to provide rich information when evaluating object detection performance.  

To conclude, our proposed methods reflect the typical ``L"-shape in LiDAR observations, which is also observed in LiDAR-based object detection networks.

\begin{table}[h]
	\caption{\label{table:L-Shape} Change of $AP_{BEV}$(\%) of two types of predicted bounding boxes compared to their original values on the KITTI \textit{val} set.}
	\centering
	\resizebox{1.00\linewidth}{!}{
	\begin{tabular}{c c c c }
		\toprule
		KITTI & Origin (Easy, Moderate, Hard) & Center aligned &  Corner aligned\\
		\cmidrule{1-4}
		STD~\cite{yang2019std}   & 90.0, 88.1, 87.7 & $+0.0, +0.1, +0.2$  & $+0.1, +0.4, +0.4$   \\
		SECOND~\cite{yan2018second} & 89.9, 87.9, 86.8 & $+0.0, +0.2, +0.1$ & $+0.1, +0.5, +0.5$     \\
		AVOD~\cite{ku2017joint}  & 88.9, 79.6, 78.9 & $-0.2, +0.0, +0.1$  & $+0.0, +6.3, +0.4$    \\
		Voxel~\cite{zhou2017voxelnet} & 81.6, 70.7, 65.8 & $+1.2, +1.1, +1.0$  & $+3.0, +4.4, +8.0$    \\
		PointRCNN~\cite{Shi_2019_CVPR} & 88.8, 86.3, 86.0 & $+0.0, +0.4, +0.7$ & $+0.2, +0.7, +1.2$ \\
		\cmidrule{1-4}
		Waymo & Origin ($<30$m, $<50$m, $<70$m)& Center aligned &  Corner aligned\\
		\cmidrule{1-4}
		PIXOR~\cite{yang2018pixor} & $62.2, 54.3, 48.5$ & $+3.8, +5.5, +4.0$  & $+4.3, +6.7, +6.9$    \\
		\bottomrule
	\end{tabular}
	}
\end{table}

\subsection{Label Quality Analysis}\label{sec:experimental_result:label_wellness}
In this section, we study how the proposed spatial uncertainty captures the quality of bounding box labels in the KITTI dataset. We evaluate the spatial uncertainty by the JIoU score between the (deterministic) object label and its spatial distribution from the our proposed generative model (``jiou-gt''). It is compared with ``cvx-hull-iou'' proposed by Meyer \textit{et al.}~\cite{meyer2019learning}, which approximates the spatial distribution by measuring the IoU value between the bounding box label of an object and its convex hull of aggregated LiDAR points. We also calculate the number of points within an object as a simple heuristic (``num-points''). All three methods range between $0$ and $1$, with larger scores indicating better label quality. 

\begin{figure}
    \centering
    \begin{minipage}{0.47\textwidth}
	\centering
	\subfigure{\label{fig:cvx_iou-npoints}\includegraphics[width=0.32\textwidth]{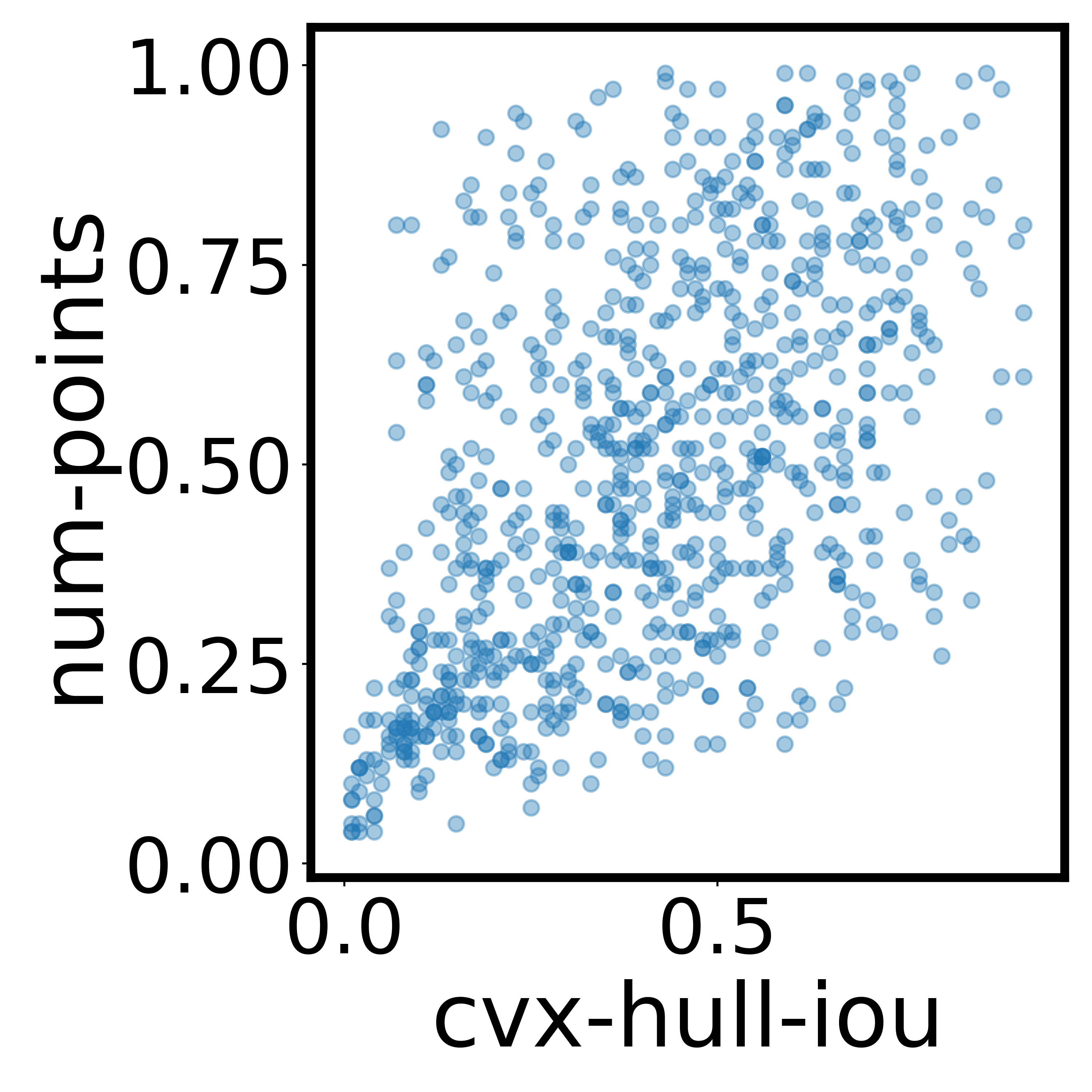}}
	\subfigure{\label{fig:cvx_iou-jiou_gg}\includegraphics[width=0.32\textwidth]{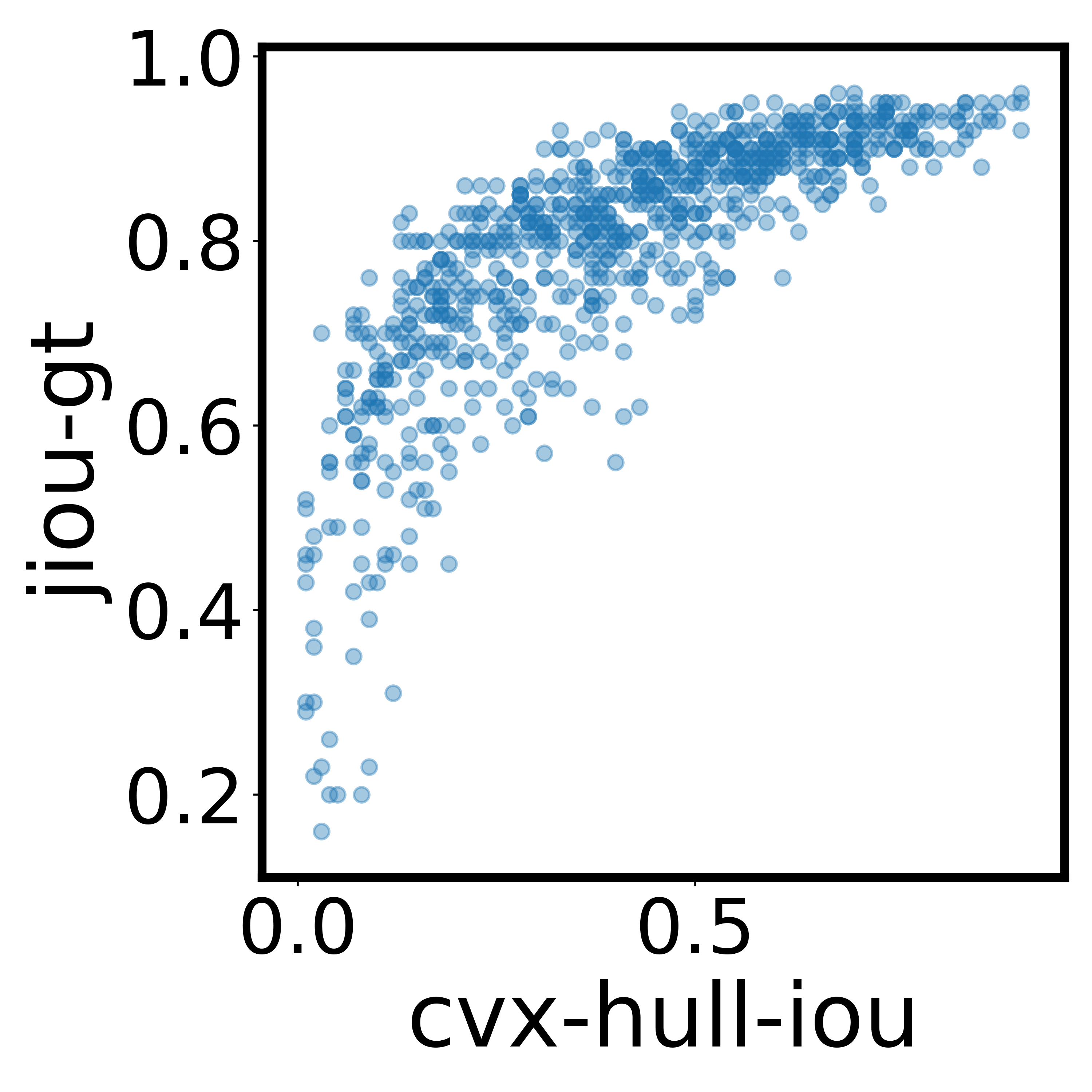}}
    \subfigure{\label{fig:npoints-jiou_gg}\includegraphics[width=0.32\textwidth]{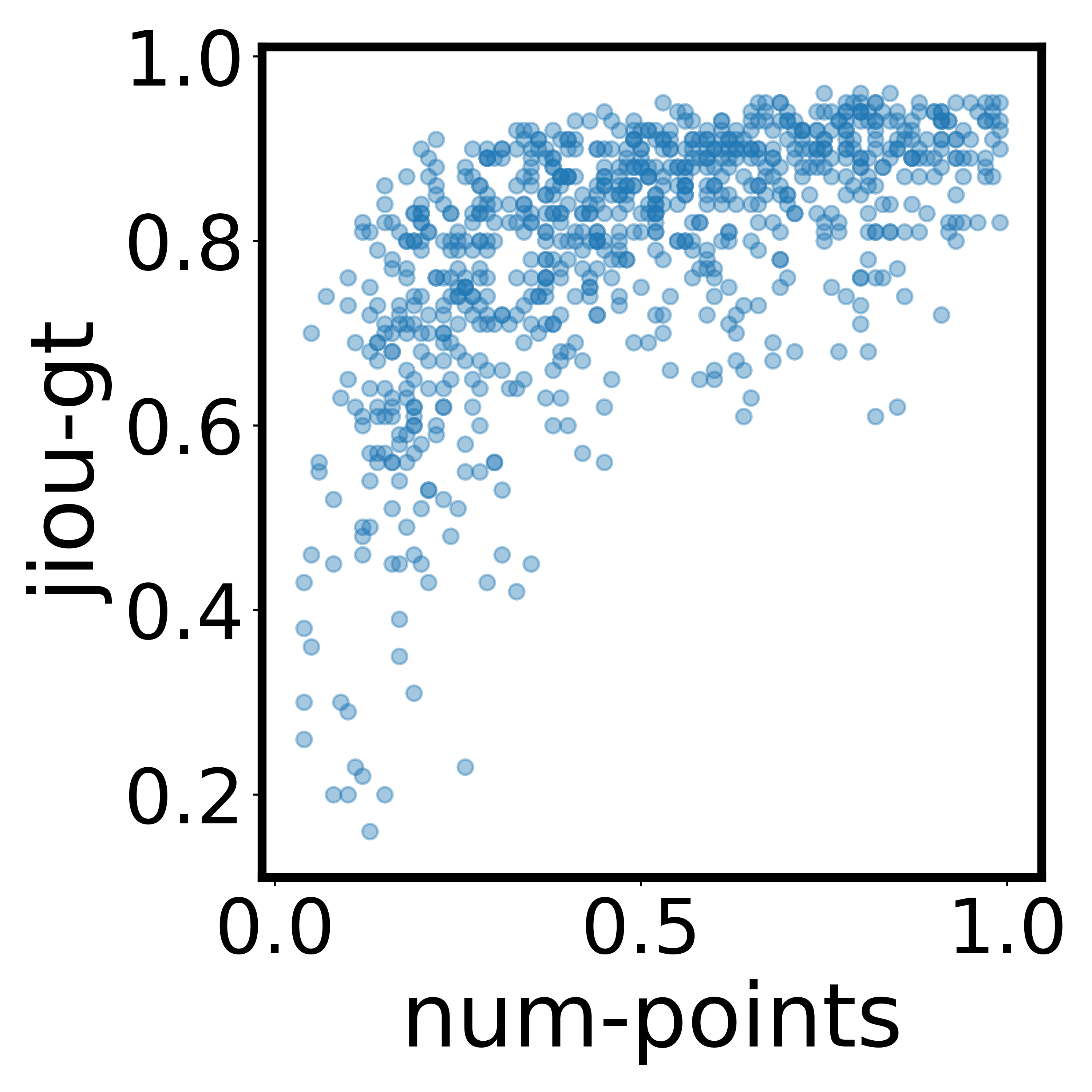}}
	\caption{Distributions of different label uncertainty measures for all objects in the KITTI \textit{val} set. Our method (``jiou-gt'') is compared with convex hull method (``cvx-hull-iou'') proposed by~\cite{meyer2019learning}, as well as the simple heuristic that calculates the number of LiDAR observations within an object (``num-points''). All three uncertainties range among $[0,1]$.}\label{fig:label_uncertainty_comparison}
\end{minipage}
\vspace{-4mm}
\end{figure}

First, we study the relationships among three methods. Fig.~\ref{fig:label_uncertainty_comparison} illustrates their distributions for all objects in the KITTI \textit{val} set. We observe that the relationship between ``cvx-hull-iou'' and ``num-points'' is relatively small (Fig.~\ref{fig:cvx_iou-npoints}). This is because they capture different aspects of spatial uncertainty. ``cvx-hull-iou'' assumes that if larger parts of an object are observed, the uncertainty is smaller.  In contrast ``num-points'' focuses on the observation density. ``jiou-gt'' is highly related both with ``cvx-hull-iou'' and ``num-points'' (Fig.~\ref{fig:cvx_iou-jiou_gg} and Fig.~\ref{fig:npoints-jiou_gg}), indicating that our proposed method naturally considers both the size and density of observed region when inferring spatial uncertainty. 

\begin{figure}[tpb]
	\centering
	\begin{minipage}{0.45\linewidth}
		\centering
		\includegraphics[width=0.8\linewidth]{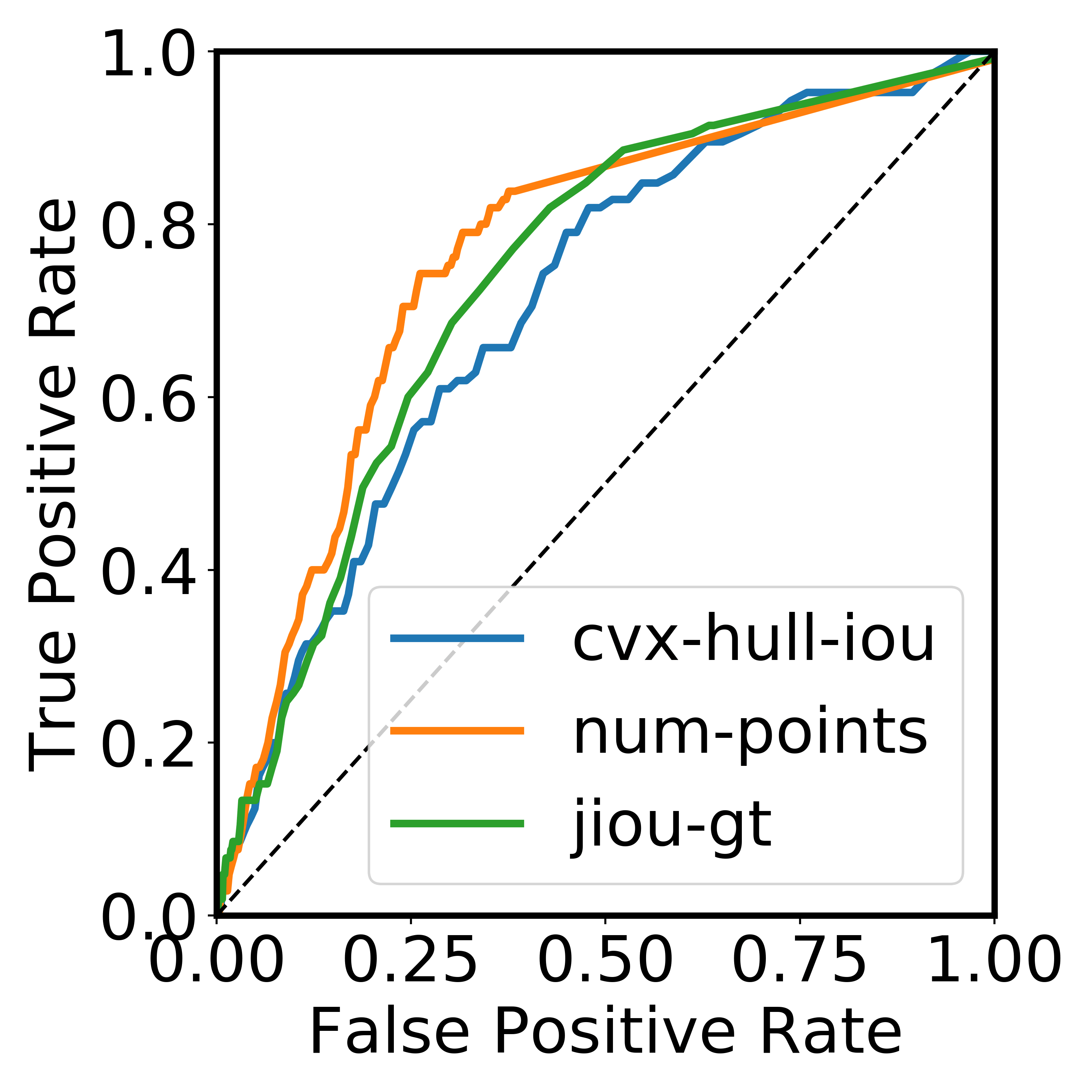}
	\end{minipage}
	\hskip 2pt
	\begin{minipage}{0.4\linewidth}
		\centering
		\caption{ROC curves for detecting bad labels in the KITTI \textit{val} set, by thresholding spatial uncertainty scores. We compare among three spatial uncertainty methods, including ``jiou-gt'', ``cvx-hull-iou'', and ``num-points''.} \label{fig:roc_curve}
	\end{minipage}
	\vspace{-3mm}
\end{figure}

Then, we study how three spatial uncertainty methods behave on objects which tend to have large label noises. 
Ideally, the modelled spatial uncertainty should depict higher values in bad labels than the labels with high quality. However, it is difficult to directly study the label noises within object detection datasets, as they do not provide ``ground truths'' of label uncertainty. Instead, we leverage the common false negatives from several state-of-the-art detectors, with the assumption that objects with bad labels are also difficult to be detected. In this regard, we collect the common false negatives from detectors listed in Tab.~\ref{table:L-Shape} as bad examples, and the rest of the objects as good labels. We threshold spatial uncertainty scores to discriminate good labels from the bad ones. From the ROC curves in Fig.~\ref{fig:roc_curve}, we observe that ``jiou-gt'' performs better than ``covx-hull-iou'', showing that it captures more reliable spatial uncertainty. ``num-points'' outperforms the other two methods, because we select bad labels based on predictions from LiDAR-based object detectors, whose classification performances highly depend on the number of LiDAR observations. We leave it as an interesting future work to do evaluation with ``ground truths'' of label uncertainty, which can be generated by querying human annotators or by simulation. 

To conclude, our proposed spatial uncertainty captures the quality of bounding box labels.

\subsection{JIoU on Different Networks}\label{sec:experimental_result:jiou}
The proposed JIoU, as an extension of IoU, allows us to evaluate the prediction or label uncertainties under the same evaluation framework of deterministic object detection. In this section, we demonstrate how to get insights of probabilistic detections with the help of JIoU. We employ a state-of-the-art probabilistic object detector called ``ProbPIXOR'' from~\cite{feng2019can}, which models data-dependent (aleatoric) uncertainty on PIXOR~\cite{yang2018pixor} - a deterministic LiDAR-based object detection network. ProbPIXOR assumes the regression variables are Gaussian-distributed with diagonal covariance matrix. Since ProbPIXOR is shown to predict over-confident or under-confident uncertainties in~\cite{feng2019can}, we further calibrate its uncertainty estimation based on temperature scaling proposed in~\cite{feng2019can} and call the new network ``CalibProbPIXOR''. The only difference between CalibProbPIXOR and ProbPIXOR is the scale of variances in the Gaussian distribution. We calculate the JIoU between the predicted and the label bounding boxes for all three networks (PIXOR, ProbPIXOR and CalibProPIXOR), as well as the ground truth JIoU following the method in Section~\ref{sec:uncertainty}. The aleatoric uncertainties of ProbPIXOR and CalibProbPIXOR are used to construct the distribution of predicted boxes, which include the box center position, length, width and heading. As for PIXOR, JIoU is directly applied to the deterministic box predictions. Note that all three network are not optimized for JIoU metric, and ProbPIXOR and CalibProbPIXOR only produce symmetric probability distributions due to the bounding box encoding. 

\begin{figure}[tpb]
	\centering
	\begin{minipage}{1\linewidth}
		\centering
		\includegraphics[width=1\linewidth]{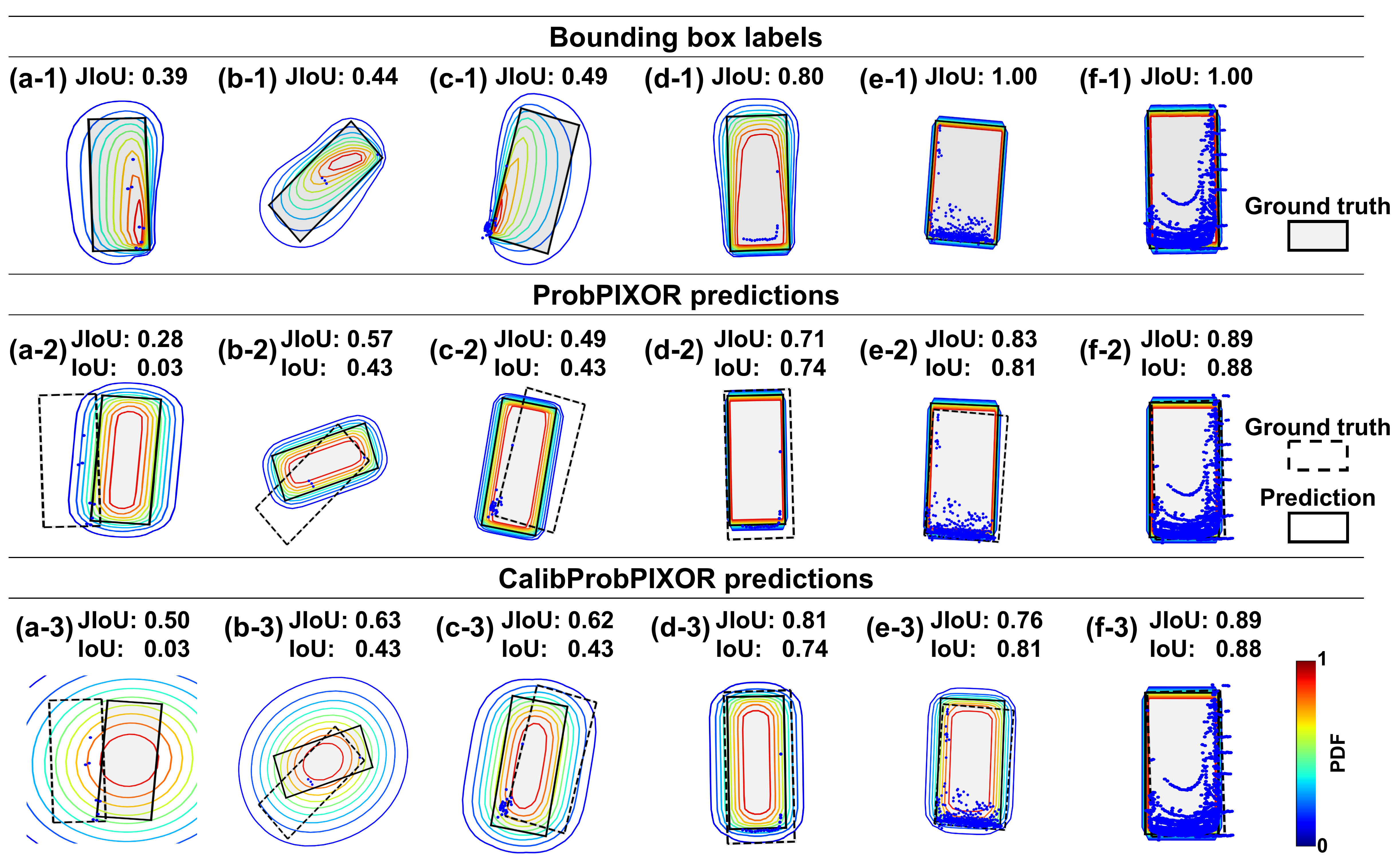}
		\vspace{-3mm}
		\caption{Some detection examples and their JIoU scores in the KITTI dataset. We visualize the spatial distribution for the bounding box labels in the first row, predictions from ProbPIXOR in the second row, and predictions from CalibProbPIXOR in the thrid row.} \label{fig:experiment_uncertainty_example}
	\end{minipage}
	\vspace{-3mm}
\end{figure}

We first explore how JIoU behaves for predictions from ProbPIXOR and CalibProbPIXOR in the KITTI dataset. We observe that JIoU scores are in general consistent with IoU scores, with higher IoU corresponding to larger JIoU, as illustrated in Fig.~\ref{fig:experiment_uncertainty_example}. However, JIoU provides us additional uncertainty information to evaluate detections. For example, Fig.~\ref{fig:experiment_uncertainty_example}(a-2) shows a bad detection with only IoU=$0.03$. However, the ground truth JIoU is already small with JIoU=$0.39$ (Fig.~\ref{fig:experiment_uncertainty_example}(a-1)) due to sparse LiDAR observations, indicating that over-emphasizing the detection performance for this object is unnecessary. Another example are detections in Fig.~\ref{fig:experiment_uncertainty_example}(b-2) and Fig.~\ref{fig:experiment_uncertainty_example}(c-2). Both are measured with the same IoU but different JIoU scores, because of different label and predictive probability distributions. Furthermore, we observe in the third row of Fig.~\ref{fig:experiment_uncertainty_example} that the predictive probability distributions become wider after calibration, because most predictions are over-confident (similar to~\cite{feng2019can}). This results in improved JIoU scores in most cases compared to ProbPIXOR. However, temperature scaling only improves the uncertainty on the whole dataset, and does not guarantee that each detection is better-calibrated~\cite{feng2019can}. Therefore, we observe a worse JIoU after calibration in Fig.~\ref{fig:experiment_uncertainty_example}(e-3). 

\begin{figure}[tpb]
	\centering
	\subfigure[On the KITTI \textit{val} set.]{\includegraphics[width=4cm]{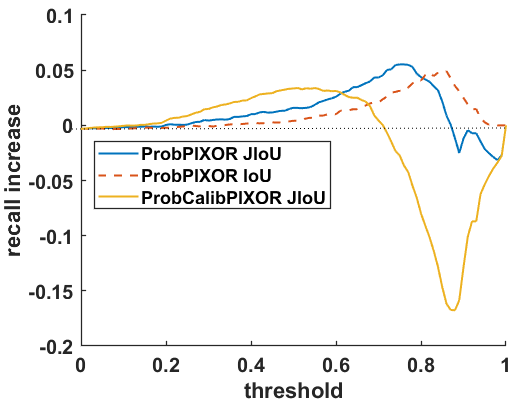}}
	\subfigure[On Waymo dataset.]{\includegraphics[width=4cm]{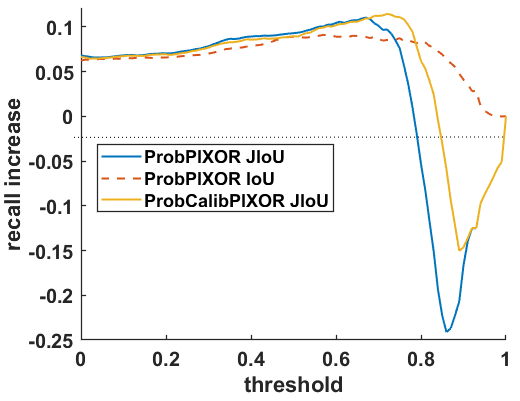}}
	\caption{Increase of recall from ProbPIXOR and CalibProbPIXOR compared with PIXOR, by thresholding detections based on IoU or JIoU.} 
	\label{fig:JIoU_exp1}
	\vspace{-5mm}
\end{figure}
Next, we use JIoU to quantitatively analyze how modelling uncertainties affects detections. Fig.~\ref{fig:JIoU_exp1} shows the relative recall gain of probabilistic object detectors (ProbPIXOR and CalibProbPIXOR) compared with the original PIXOR network, by thresholding detections based on IoU or JIoU metrics. In both KITTI and Waymo datasets, we observe consistent recall improvement by modelling uncertainty based on the IoU metric (see ``ProbPIXOR IoU'' in Fig.~\ref{fig:JIoU_exp1}), indicating higher Average Precision performance (as shown in~\cite{feng2018leveraging}). When using JIoU, however, both ProbPIXOR and ProbCalibPIXOR do not perform well at high threshold (e.g. JIoU$>0.8$), because they only produce symmetric probability distribution (as discussed above), and cannot well calibrate our proposed spatial distribution. The result shows that JIoU effectively penalize the distribution mismatch between two probabilistic boxes especially at high values, and indicates the potential improvements for probabilistic modelling in object detection networks (e.g. better bounding box encodings or assuming correlation between regression variables~\cite{harakeh2019bayesod}).   

To conclude, JIoU provides us richer information than IoU to evaluate probabilistic object detection networks.
\section{Discussion and Conclusion} \label{sec:conclusion}

In this work, we propose a generative model to infer the uncertainty inherent in bounding box labels of object detection datasets with LiDAR point clouds. The label uncertainty includes object shape and measurement information, and can represent non-diagonal correlation of label parameters. We further propose a new spatial distribution to visualize and represent the uncertainty of 2D or 3D bounding boxes. Finally, we propose JIoU, as an extension of IoU, to evaluate probabilistic object detection. Comprehensive experiments on KITTI and Waymo datasets verify our proposed method.

Our work can be extended in several ways. For example, it is possible to incorporate the proposed label uncertainties to train an object detector, in order to improve its robustness against noisy data. Using JIoU to evaluate uncertainty estimation performance among different probabilistic object detectors (e.g.~\cite{feng2018leveraging,harakeh2019bayesod}) is another interesting future work.  






\bibliographystyle{IEEEtran}
\bibliography{bibliography}

\end{document}